# Discovering Cyclists' Visual Preferences Through Shared Bike Trajectories and Street View Images Using Inverse Reinforcement Learning

Kezhou Ren [a,b], Meihan Jin [c], Huiming Liu [d], Yongxi Gong [a,b], Yu Liu [e]

[a] School of Architecture, Harbin Institute of Technology Shenzhen, Shenzhen 518055, PR China

[b] Key Research Base of Humanities and Social Sciences of Guangdong Province, Center for Digital Technology of Space Governance, Harbin Institute of Technology Shenzhen, Shenzhen 518055, China

[c] College of Urban Transportation and Logistics, Shenzhen Technology University, Shenzhen 518118, PR China

[d] Faculty of Humanities and Arts, Macau University of Science and Technology, Macau 999078, China

[e] Institute of Remote Sensing and Geographical Information Systems, School of Earth and Space Sciences, Peking University, Beijing 100871, PR China

## Author List:

*Corresponding author: Yongxi Gong, E-mail: yongxi_gong@163.com

Author: Kezhou Ren, E-mail: 1173450209@stu.hit.edu.cn

Author: Meihan Jin, E-mail: jmeihan@hotmail.com

Author: Huiming Liu, E-mail: hmliu@must.edu.mo

Author: Yu Liu, E-mail: liuyu@urban.pku.edu.cn

# Abstract

Cycling has gained global popularity for its health benefits and positive urban impacts. To effectively promote cycling, early studies have extensively investigated the relationship between cycling behaviors and environmental factors, especially cyclists' preferences when making route decisions. However, these studies often struggle to comprehensively describe detailed cycling procedures at a large scale due to data limitations, and they tend to overlook the complex nature of cyclists' preferences. To address these issues, we propose a novel framework aimed to quantify and interpret cyclists' complicated visual preferences by leveraging maximum entropy deep inverse reinforcement learning (MEDIRL) and explainable artificial intelligence (XAI). Implemented in Bantian Sub-district, Shenzhen, we adapt MEDIRL model for efficient estimation of cycling reward function by integrating dockless-bike-sharing (DBS) trajectory and street view images (SVIs), which serves as a representation of cyclists' preferences for street visual environments during routing. In addition, we demonstrate the feasibility and reliability of MEDIRL in discovering cyclists' visual preferences. We find that cyclists focus on specific street visual elements when making route decisions, which can be summarized as their attention to safety, street enclosure, and cycling comfort. Further analysis reveals the complex nonlinear effects of street visual elements on cyclists' preferences, offering a cost-effective perspective on streetscape design. Our proposed framework advances the understanding of individual cycling behaviors and provides actionable insights for urban planners to design bicycle-friendly streetscapes that prioritize cyclists' preferences.

# 1 Introduction

Cycling, widely recognized as a sustainable means of transportation, promotes outdoor activities and presents a potential solution to urban challenges such as traffic congestion and air pollution (Cai et al., 2023; Lazarus et al., 2020; Song et al., 2024; Wang et al., 2024). The emergence of dockless bike sharing (DBS) has further enhanced cycling by significantly improving accessibility to active travel, seamlessly integrating with urban public transportation systems. Therefore, DBS has garnered widespread adoption across numerous countries, sparking remarkable academic interest in transportation and urban planning (Albuquerque et al., 2021; Caigang et al., 2022; Ding et al., 2022). These interests are mainly focused on the complexity of cyclist behaviors, the interplay with public transit, and the use of cycling infrastructure to promote the use of bicycles (Aziz et al., 2018; Bao et al., 2017; Zare et al., 2022).

**As a planning strategy to substantially promote cycling** (Bai et al., 2022), **built environment is recognized to have crucial impacts on bicycle usage from several aspects** (El-Assi et al., 2017; Liu et al., 2024; Xu et al., 2019). Some studies assessed how environmental factors influence individual cycling willingness (Liu et al., 2024; Morton et al., 2021; Wang et al., 2024; Zhao et al., 2019), while others explored the relationship between built environments and trip volume, thereby predicting bicycle demand at a regional level (Gao et al., 2023; Li et al., 2021; Xu et al., 2019). Notably, some studies have highlighted the role of bicycles in urban public transportation (Liu et al., 2024; Zhang et al., 2024; Zhou et al., 2019), specifically how built environments influence the integration of DBS with metro at both station (Cheng et al., 2022; Guo & He, 2020) and origin-destination (OD) levels (Fu et al., 2023; Guo & He, 2021). These studies provide valuable insights into cyclists' mode selection and inform effective planning for the connection between different means of transportation.

**However, most works focus primarily on OD, rather than the effect of built environment on biking behavior in the detailed cycling procedures.** Considering that cyclists have direct contact with the street environment without any physical isolation, and will make cycling decisions in real-time based on the physical environment of the street (Wang et al., 2024; Zare et al., 2024; Zhou et al., 2024), it is crucial to understand cycling procedure between places under the impacts of surroundings (Guo & He, 2020, 2021; Zare et al., 2024). One of the most common and intuitive frameworks is Route Choice Modeling (RCM), which details the process of individual route selection and predicts the paths they are likely to choose (Prato, 2009). Typically, current research focuses on behaviors rather than motivations (Zare et al., 2024), with the general behavioral patterns identified from individual cycling procedures referred to as preferences (Prato, 2009; Stinson & Bhat, 2003; Zhao & Liang, 2023). Early investigations integrate RCM and questionnaires to discern cyclists' preferences based on street environment characteristics (Stinson & Bhat, 2003). However, survey results may not entirely reflect cyclists' actual choice due to the absence of real riding conditions, which may lead to biased estimation of cyclists' preferences. Recent studies have highlighted the potential of integrating RCM with trajectory data (Bao et al., 2017; Hu, 2024; Huang et al., 2021). By incorporating detailed individual route decision information and diverse route attributes, these studies offer insights into cyclists' environmental preferences (Gupta & Gunukula, 2024; Koch & Dugundji, 2020). Nevertheless, current research has predominantly focused on macro-level environmental factors within RCM, neglecting the cyclists' visual

experiences (Blitz, 2021; Guo & He, 2021; Mertens et al., 2016; Porter, 2018; Zare et al., 2024). While some studies have attempted to explore the impact of streetscape characteristics from cyclists' perspectives (Guan et al., 2023; Huang et al., 2014; Jeon & Woo, 2024; Song et al., 2024), they frequently assume linear utility functions, overlooking the non-linear nature of cyclists' environmental preferences. Therefore, we need a more flexible framework to discover the complicated street visual preferences of cyclists based on their detailed route decision procedures.

In recent years, reinforcement learning (RL) has proven highly effective for sequential decision-making in dynamic environments (Li, 2023; Qin et al., 2022) and has been widely used in route recommendation, vehicle repositioning (Rong et al., 2016; Verma et al., 2017; Yu & Gao, 2022), and other applications (Gao et al., 2018; Luo et al., 2022). The similarity between RL's training process and route selection highlights its significant potential in trajectory data mining. RL methods often rely on predefined operating rules. In contrast, inverse reinforcement learning (IRL) observes the behaviors of navigators as they move between ODs, and then recovers the routing principles they likely follow through a combination of simulation and optimization (Abbeel & Ng, 2004; Alger, 2016; Fahad et al., 2018). IRL has been widely employed to discover the complex routing tendency of taxi drivers (Liu et al., 2023; Zhao & Liang, 2023). It is also well suited in discovering cyclists' preferences based on trajectory data due to its structural similarity to RCM (Koch & Dugundji, 2021; Rust, 1987; Safarzadeh & Wang, 2024; Zhao & Liang, 2023). Meanwhile, the flexibility of IRL provides researchers with opportunities to comprehensively understand the detailed procedures of cycling.

In this study, we propose a data-driven framework that integrates multi-source big data and IRL to quantify and interpret the cyclists' visual preferences as they travel from their initial to goal positions. Specifically, we take Bantian Sub-district in Longgang District, Shenzhen as the empirical research area. By applying Maximum Entropy Deep Inverse Reinforcement Learning (MEDIRL), we discover cyclists' general visual preferences by integrating DBS trajectories and street view images (SVIs) from a more bottom-up perspective. We then validate the framework by simulating trajectories based on these preferences and comparing them to actual cyclist trajectories. Finally, we adopt XAI techniques to illustrate the characteristics of streetscapes that cyclists prefer.

The rest of the paper is organized as follows. Section 2 introduces the study area and dataset. Section 3 outlines our methodological framework, including MEDIRL, XAI and their adaptability. In section 4, we empirically validate the reliability and explainability of our approach within the study area. Section 5 summarizes this research and outlines the future work.

## 2 Study Area and Dataset

### 2.1 Study Area

Our framework is mainly applied as a case study in Bantian Sub-district of Longgang District, Shenzhen (shown in **Fig. 1**). This area is characterized by relatively consistent cycling demand, convenient traffic facilities, comfortable travel conditions and diverse environmental attributes. Therefore, it is suitable for investigating cyclists' preferences to environmental factors.

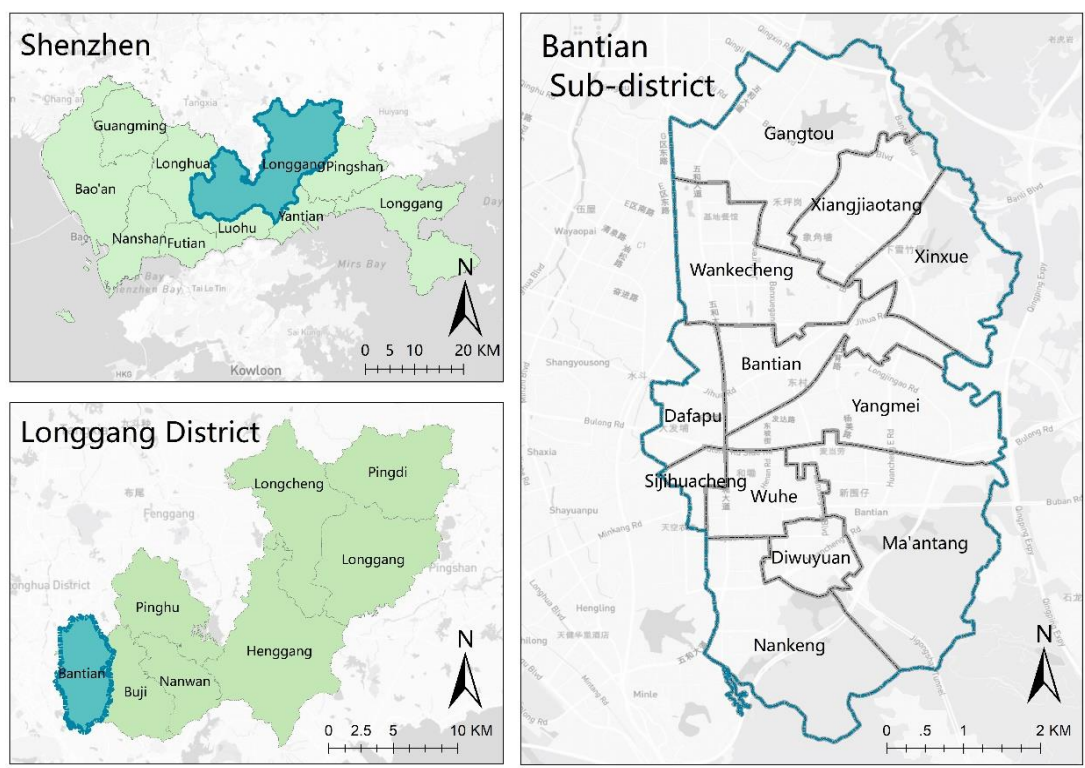

**Fig. 1. The location of Bantian Sub-district**

Bantian Sub-district has served as a showcase for the integration of technology and urban development within the Guangdong-Hong Kong-Macao Greater Bay Area. It is particularly renowned for its technology industry, and attracts a large number of young commuters. Situated on the border of three districts in Shenzhen (Futian, Longgang, Longhua), the area offers diverse transportation options for residents. Specifically, it boasts well-established public transportation, including 2 metro lines with 5 subway stations, complemented by a sufficient supply of DBS services. The total road network spans 38.97 km, comprising 5,229 intersections and 7,039 road segments accessible to cyclists. Moreover, Bantian Sub-district provides favorable conditions for year-round outdoor cycling with an average temperature of 23.3 ℃ and gentle slopes averaging less than 10°. Its spatial heterogeneity in natural and socio-economic environments enables us to identify the cycling behavior affected by the environment and extract typical attributes.

## 2.2 Dataset

### 2.2.1 DBS Trajectory Data

DBS trajectory data provide crucial insights into individual route choices, which are essential for uncovering cyclists' behavioral patterns. The specific data utilized in this study are sourced from a well-known DBS service provider collected from Nov 1st to Nov 30th, 2017. Specifically, our DBS trajectory data encompass basic order details such as Order ID, User ID, DBS ID, Start Time, End Time, Start Coordinate, End Coordinate and a sequence of trajectory points collected at three-second intervals, as illustrated in **Table 1**.

**Table 1** Sample of a DSB trajectory record.

| Attribute | Value Example |
|---|---|
| Order ID | 1628190 |
| Shared Bike ID | AA659656D9F8BD8F7B20******** |
| Start Time | 2017-11-07 08:16:41.0 |
| Start Coordinates | 114.26739014, 22.70951389 |
| End Time | 2017-11-07 08:31:07.0 |
| Start Coordinates | 114.24780835, 22.70009927 |
| Trajectory Point | 114.267402,22.709543;1510013805840#114.267402,22.709543; |

For further analysis, we process the DBS trajectory data using the following steps. First, we identify and eliminate the counter-intuitive cycling trips in order to enhance data quality, and obtain a dataset of 21187 distinct cycling trips suitable for further processing (Wang et al., 2024). Second, we apply a map-matching method based on Hidden Markov Model (HMM) to rectify the trajectory points on the road network (Meert & Verbeke, 2018). Thus, the GPS trajectory points are converted to sequences of road segments. Additionally, to mitigate potential biases in estimating preferences due to temporal changes, we focus our analysis on cyclists' behaviors during weekdays and daylight hours. Furthermore, preliminary experiments indicate that approximately 83.8% of cyclists select the shortest path for trips involving fewer than five road segments, indicating that environmental factors may have limited explanatory power. Therefore, we concentrate on cycling trips with more than five road segments, resulting in 15052 distinct trip trajectories as the valid dataset for further investigation. The results of exploratory data analysis are presented in Appendix.

### 2.2.2 SVI

Cyclists often lack visibility of the destination and attributes of entire routes (Huang et al., 2014). Consequently, they tend to make route choices based on their on-situ surroundings (Guan et al., 2023; Huang et al., 2014; Song et al., 2024; Zare et al., 2024). SVI, a widely used type of big geospatial data, offers detailed visual representations of urban physical environments (Ito & Biljecki, 2021; Zhang et al., 2018). Additionally, SVIs implicitly capture invisible information about socio-economic conditions (Zhang et al., 2019, 2019, 2021). As a result, SVIs provide a more comprehensive understanding of cycling environments and are well-suitable for simulating cyclists' real-time visual perceptions on urban streets.

The dataset we used in this study is obtained from Baidu, comprising 7924 distinct panoramas in Bantian Sub-district and dated between 2016 to 2018. Semantic segmentation is conducted using the SegNet (Badrinarayanan et al., 2016) model pre-trained on the Cityscapes dataset (Cordts et al., 2016), to extract visual elements. As a result, we identify eight primary categories and 21 detailed subcategories of street scenery, as detailed in **Table 2**. Subsequently, we quantify the cyclists' visual perceptions by calculating the ratios of the pixels assigned to each category relative to the total number of pixels in the image, normalized using min-max scaling between zero and one. The segmentation achieves an average coverage of 97.88%, indicating its applicability for further investigation.

**Table 2** Classification Standard for Cityscapes Dataset

| Primary Categories | Sub Categories | Definition |
| --- | --- | --- |
| Flat | Road | Part of the ground on which cars usually drive. |
| | Sideway | Part of the ground designated for pedestrians or cyclists. |
| Human | Pedestrian | A human that is walking. |
| | Cyclist | A human that would use some device to move |
| Vehicle | Car | Car, jeep, SUV, van with continuous body shape, caravan. |
| | Truck | Truck, box truck, pickup truck. |
| | Bus | Bus, public transport or long distance transport. |
| | Train | Vehicle on rails. |
| | Motorcycle | Motorbike, moped, scooter. |
| | Bicycle | Bicycle without the driver. |
| Construction | Building | Building, skyscraper, house, bus stop building, car port. |
| | Wall | Individual standing wall. |
| | Fence | Fence including any holes. |
| Object | Pole | Small mainly vertically oriented pole. |
| | Traffic Light | The traffic light box without its poles. |
| | Traffic Sign | Traffic- signs, parking signs, direction signs - without their poles. |
| Nature | Vegetation | Tree, hedge, all kinds of vertical vegetation. |
| | Terrain | Grass, all kinds of horizontal vegetation, soil or sand. |
| | Sky | Open sky, without leaves of tree. |

## 3 Methodology

### 3.1 Framework

In our study, we view the cycling procedure as a continuous decision-making journey driven by cyclists' visual preferences within the constraints of their ODs. Our goal is to discover cyclists' general visual preferences influenced by streetscape characteristics derived from this procedure. This is challenging due to the complexity of their on-situ environments, the stochastic nature of cyclists' behaviors influenced by surroundings, and the abundant spatio-temporal and semantic information involved in cycling.

In response, we propose an IRL-based framework to quantify and interpret cyclists' visual preferences along urban streets when they travel from origins to destinations, focusing specifically on their cycling procedures. The conceptual research diagram is described in **Fig 2**. IRL is ideal for our research scenario due to its efficiency in mining sequential dependencies and semantic information, as well as its flexibility in integrating DL architectures and high dimensional features (Liu et al., 2020, 2023; Wulfmeier et al., 2015, 2017), which helps discover complicated relationships. Furthermore, IRL's unique training process makes it more behaviorally interpretable compared to conventional DL methods and facilitates further simulation and optimization (Zhao & Liang, 2023).

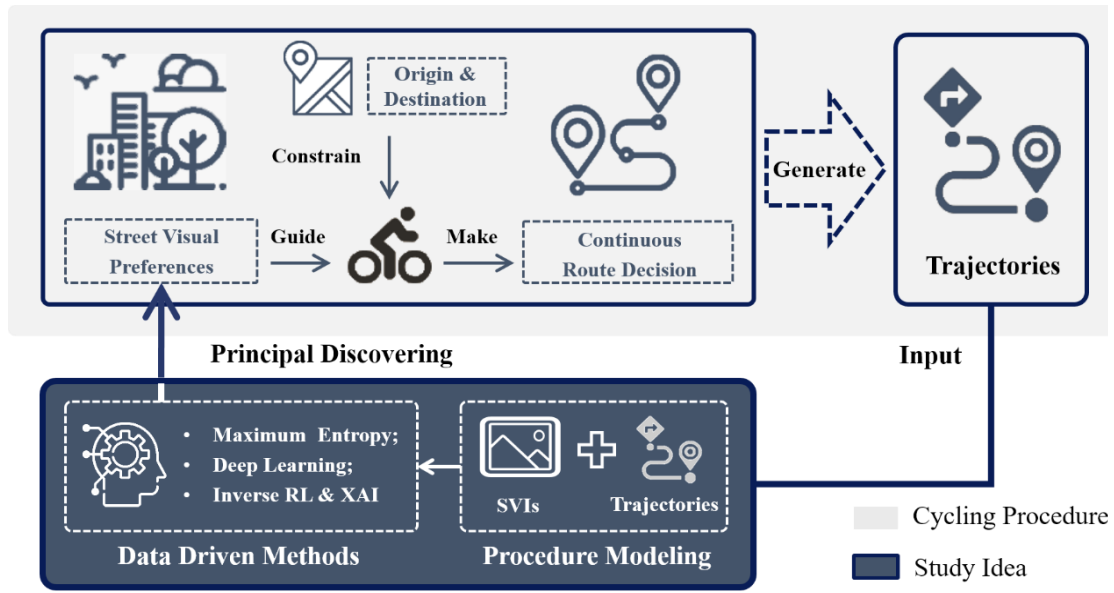

**Fig. 2. Conceptual research diagram**

The overall workflow comprises three distinct steps, as illustrated in **Fig. 3**. First, cycling is treated as a route decision process constrained by road spatial networks, taking into account origin-destination (OD) pairs and cyclists' on-situ environments. We formalize it as a Markov Decision Process (MDP) by integrating SVIs and DBS trajectories to detail cycling procedures outlined earlier for further analysis.

Second, we employ IRL to recover the underlying reward function of MDP from observed trajectory data. The reward function reflects the general principal cyclists follow, influenced by cycling environments, and serves as a quantified representation of cyclists' street visual preferences. We approximate this reward function using a combination of maximum entropy model and deep neural network (DNN) to balance diverse cyclist preferences and capture their non-linear nature. In summary, MEDIRL identifies cyclists' street visual preferences derived from the cycling procedures through a data-driven approach (Gupta & Gunukula, 2024; Safarzadeh & Wang, 2024; Wulfmeier et al., 2015, 2017). To validate the performance of our model, we generate trajectories for each trip based on their OD pairs. We then compare the similarities between real and synthetic trajectories at both the statistical and trip levels.

Finally, we utilize XAI to unravel the contributions of visual elements extracted from SVIs. Since SVIs provide a comprehensive understanding of cycling environments, preferences for particular street visual elements may reflect cyclists' potential demands. Therefore, we further interpret these demands in light of the earlier analyses.

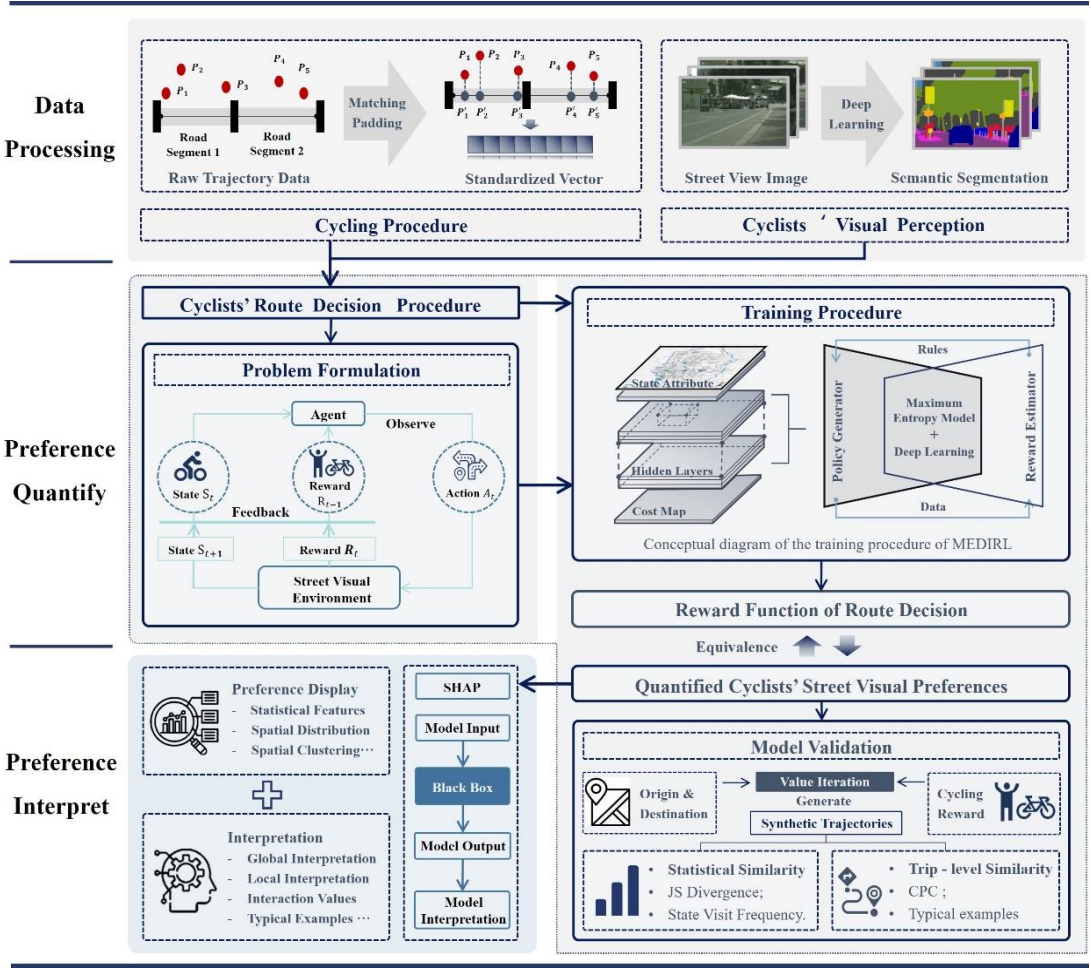

**Fig. 3. Overall workflow of the research**

### 3.2 Preliminaries and Problem Formulation

The similarity between RL's training process and route selection highlights its significant potential in trajectory data mining. Generally, navigators are viewed as agents, and traditional RCM problems can be mathematically formulated as an MDP. This process serves as the basis for implementing IRL (Alger, 2016; Liu et al., 2020; Wulfmeier et al., 2015, 2017). An MDP is generally defined as $M = \{S, A, T, R, \gamma\}$, where $S$ denotes the state space, representing the set of possible positions the agents can be; $A$ denotes the set of possible actions the agent can take. $T(s_t, a_t, s_{t+1})$ denotes a transition model that determines the next state $s_{t+1}$ given the current state $s_t$ and action $a_t$. $R(s, a)$ is the reward function, defined as the feedback obtained by the agent when taking action $a \in A$ in state $s \in S$. It is also the principal that cyclists follow in the RL training process. $\gamma$ is the discount rate, and it controls how much future rewards are valued compared to immediate rewards, allowing agents make decisions based not only on their real-time perceptions but also on their understanding of entire routes.

In the modeling of sequential decision-making problems, the policy $\pi(a|s)$ describes the moving strategy at each state. The agent's policy is often non-deterministic, and this stochastic policy can be intuitively understood as the probability of the agent taking action $a_t \in A$ given the current state $s_t \in S$. A trajectory represents the sequence of states, actions and corresponding rewards experienced by the agent under the policy $\pi$ for a specific OD pair. It is defined as $\{(s_1, a_1, r_1), (s_2, a_2, r_2), \cdots, (s_t, a_t, r_t)\}$, with each tuple denoting a state-action-reward triplet. In the traditional RL setting, the reward function $R(s, a)$ $s \in S, a \in A$ is predefined. The objective is to find the optimal policy $\pi^*$ that maximizes the expected cumulative reward over trajectories.

In our study, cyclists are modeled as MDP agents navigating within a discretized grid environment constrained by road spatial networks as illustrated in **Fig. 4**. Solving this MDP model will give us the optimal decision strategy for each different location.

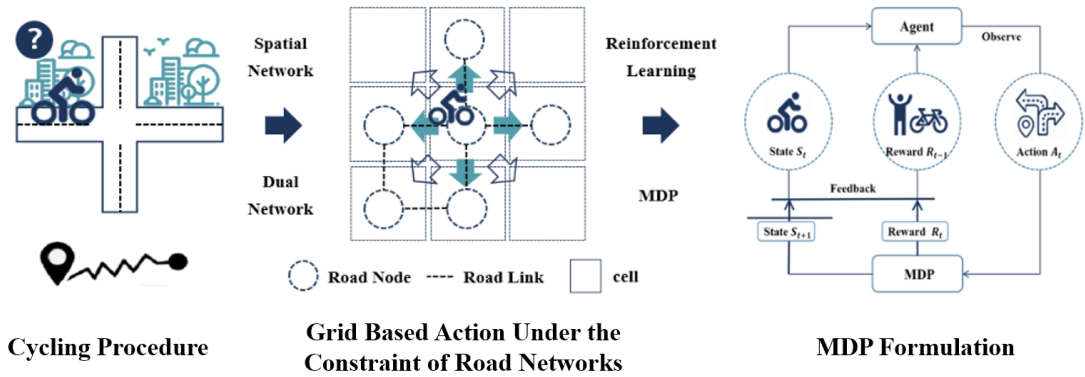

**Fig. 4. Process of problem formulation**

Specifically, we can define the key elements of the MDP in our research scenario as follows:

- **State:** Each state $s \in S$ is a vector used to describe the basis of a cyclist's decision-making, denoted as $s = \{Pos, SE\}$. The vector $Pos$ is a two-dimensional vector utilized to represent the road segments and the 100m grid units where cyclists are located. In other words, the location of cyclists is determined by both road segments and their corresponding grid. The vector $SE$ is a 23-dimensional representation of the proportions of semantic elements in SVIs, capturing the street visual environments experienced by cyclists at their locations (Liu et al., 2020; Ziebart et al., 2008).
- **Action:** An action $a \in A$ indicates the grid-to-grid movement choice under the restriction of road networks. Inspired by existing studies, we define a global action space $A$ consisting of 9 movement directions — forward (F), forward left (FL), left (L), backward left (BL), backward (B), backward right (BR), right (R), forward right (FR), and stay(ST), as shown in [Fig. 4](). Note that, although these 9 directions represent a comprehensive set of all potential actions to be taken anywhere, only a subset of them are applicable for most states. In order to account for the specific layout of the local network, we have also defined a local action space $A_s \in A$ to capture all valid actions at each location $Pos$ in each state $s$.

- **Policy:** The policy $\pi(a|s)$ describes how cyclists make route decisions under the influence of the street visual environment as they travel between ODs. The optimal policy, denoted as $\pi^*$, represents the most representative route decision pattern for cyclists. Agents following $\pi^*$ intends to maximize their cumulative reward during their own trips.

- **Reward function:** The reward function $R(s, a)$ characterizes cyclists' preferences for the street visual environment as they travel between their initial and goal positions. We use a set of parameters $\theta$ to approximate the feedback for specific actions at each state, thereby replacing the location-based reward representation with a mapping between states, actions, and their associated rewards.

In summary, our study models the cycling procedure as an MDP for further analysis. Specifically, cyclists make continuous street selection decisions based on street visual surroundings, given their ODs, aiming to maximize their cumulative reward. This is accomplished by sampling trajectory data through continuous interaction between the agent and their on-situ environment, reflecting real-world cycling procedures and fitting well within RL settings.

However, the feedback cyclists gain from streetscapes is complicated, making it challenging to manually craft a reward function that captures all the behaviors exhibited by cyclists. By reversing the RL process, IRL extracts the reward function $R_\theta(s, a)$ of MDP from demonstrated data, which is assumed to be sampled from optimal policy $\pi^*$ (Ng & Russell, 2000). The reward function can be tailored to cyclists' general street visual preferences when they are routing. As a result, our methodological approach provides a solid foundation for discovering cyclists' general street visual preferences based on their continuous route decision procedures influenced by streetscape characteristics.

### 3.3 Adapting MEDIRL for Approximating Street Visual Preferences

We introduce an MEDIRL architecture to recover non-linear reward function from complex and diverse individual cycling procedures, which is helpful to quantify cyclists' street visual preferences as they travel between ODs. We begin by describing the formulation of Maximum Entropy Inverse Reinforcement Learning (MaxEnt IRL). Subsequently, we present the adapted version, MEDIRL, along with its specific architecture designed for unraveling cyclists' street visual preferences along road networks.

Original IRL offers a valuable approach to learning reward function from trajectory data, serving as general operating rules of agents. Specifically, the IRL algorithm attempts to infer the reward function $R_\theta(s, a)$ from observed trajectories: $\tau^{(i)} = \{(s_1^{(i)}, a_1^{(i)}), (s_2^{(i)}, a_2^{(i)}), \cdots, (s_n^{(i)}, a_n^{(i)})\}$, where $\tau^{(i)}$ represents the $i-th$ trajectory, and $n$ is the length of the trajectory. However, in real-world scenarios, preferences of cyclists may vary, leading to diverse route decision patterns within trajectories. Thus, Ziebart et al. (2008) introduced the principle of maximum entropy into IRL to tackle issues of sub-optimality and randomness by manipulating the distribution over possible trajectories. According to the formula of MaxEnt IRL, the probability of observing any given trajectory $\tau$ is directly proportional to the exponential of its cumulative reward:

$$P_\theta(\tau) = \frac{1}{Z} exp(R_\theta(\tau))$$

where the partition function $Z$ is the integral of $R_\theta(\tau)$ over all possible trajectories. Therefore, we can frame the IRL problem as solving the maximum likelihood problem based on the observed trajectories:

$$\arg \max_\theta \sum_{i=1}^{N} \log(P_\theta(\tau_i))$$

However, original MaxEnt IRL typically represents the interaction between agents and the environment using linear functions, which lacks the capacity to deal with diverse and complicated street visual preferences during cycling. Therefore, our study adopts MEDIRL (Wulfmeier et al., 2015, 2017) to recover the underlying reward function and quantify cyclists' street visual preferences from DBS trajectory data. Specifically, the MEDIRL framework of this study mainly consists of two parts: policy generator and reward estimator, as illustrated in **Fig. 5**. In the subsequent sections, we present more details on each module of the algorithm.

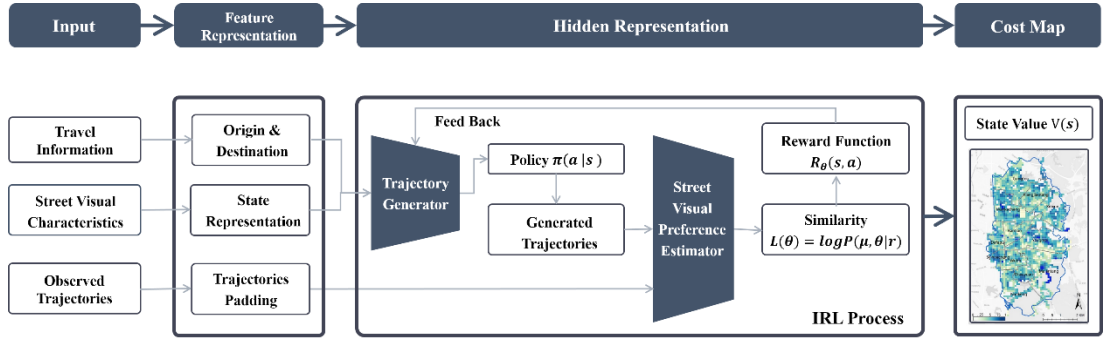

**Fig. 5. Training framework**

### 3.3.1 Trajectory Generator

The purpose of trajectory generator is to infer the distribution of state visit frequency (SVF) and the target policy $\pi_G$. In each step, we formulate a new MDP to model decision-making for DBS route choices, using the reward function estimated from the previous step. We then use a value iteration algorithm (Fahad et al., 2018; Rong et al., 2016; Yu et al., 2019) based on dynamic planning to iteratively determine the cyclist's optimal action in each state, maximizing the total cumulative rewards for a single trip. In other words, value iteration algorithm ultimately produces the optimal policy $\pi_G$ based on the current rewards. Subsequently, given a specific OD pair, our trajectory generator utilizes the learned policy to reconstruct the cyclist's trajectory, thereby estimating the expected SVF distribution under the current policy. The expectation can be expressed as:

$$E[\mu] = \sum_{\tau:\{s,a\}\in\gamma} P(\tau|r)$$

where $E[\mu]$ represents the distribution of SVF by the cyclist who follows the current policy $\pi_G$. $r$ denotes the rewards estimated in the previous step, and $\tau$ is the trajectory obtained by sampling according to the current policy given a specific OD pair.

### 3.3.2 Street Visual Preferences Estimator

The purpose of street visual preferences estimator is to infer the reward function of the optimal policy $\pi_G$ estimated by trajectory generator and then calculate the values for each state. The reward function can be interpreted as cyclists' street visual preferences (Koch & Dugundji, 2021; Rust, 1987; Safarzadeh & Wang, 2024; Zhao & Liang, 2023). In each iteration, the estimator receives input comprising real-time perceptual information, which includes the cyclist's location awareness along with corresponding environmental details. Additionally, it takes in the estimated policy outcomes $\pi_G$ and the DBS trajectory data $\mu$. Before feeding the observed trajectory data into the reward estimator, we use zero-padding to extend empty states (i.e., states with all attributes set to zero) in each trajectory until all trajectories are as long as the longest one in the dataset. Consistent with previous workings, the padded states are masked so they do not impact the estimation results. Then, we utilize the framework of MEDIRL to estimate the cyclists' general street visual environment and location preferences. The task of solving the IRL problem can be formulated as Maximum A Posteriori (MAP) estimation in the context of Bayesian inference, that is, maximizing the joint posterior distribution of observing data $\mu$ and model parameters $\theta$ under the given reward structure $r$. The loss function is expressed as follows:

$$L(\theta) = logP(\mu, \theta|r) = logP(\mu|r) + logP(\theta)$$
$$\theta: Network\ Parameters$$

where $L(\theta)$ is the loss function of MEDIRL, $P(D|r)$ represents the distribution of demonstrated data given a certain reward. By utilizing gradient descent to solve the loss function mentioned above, it becomes feasible to approximate the cyclists' street visual preferences function $R_\theta$ and their quantified preferences in each state $R = \{r_1, r_2, \cdots, r_s\}, s \in S$.

### 3.3.3 Metrics for Assessing IRL Performance

To validate our model's performance, we generate synthetic trajectories by utilizing our learned reward function and policy, which is similar to the process in a policy generator. Specifically, this procedure involves an agent-based simulation, where the optimal action in each state acts as behavioral rules for generating new trajectories. These synthetic data are expected to closely resemble real trajectories. We then employ Jensen-Shannon Divergence (JSD) to measure the similarity of statistical distributions, and Common Part of Commuters (CPC) to evaluate the similarity of individual trajectories. The detailed information of the metrics is shown below.

- **Jensen-Shannon Divergence (JSD)**: In probability theory, the JSD overcomes the asymmetry of Kullback–Leibler (KL) Divergence to measure the similarity between two probability distributions (Liu et al., 2023), and has been widely used in related topics (Rao et al., 2020; Zhao & Liang, 2023). Generally, a smaller JSD indicates a greater similarity between the distributions. In this study, we represent the probability distribution by utilizing the frequency of occurrence of each state in the trajectories. For the given observed trajectory $\tau_D$ and predicted trajectory $\tau_P$, the KL Divergence is defined as follows:

$$D_{KL}(p||q) = \sum_i p_i log \frac{p_i}{q_i}$$

Generally, $D_{KL}(p||q)$ is also referred to as the relative entropy of probability distributions $p$ and $q$. The JSD between them is defined as follows:

$$D_{Jensen-Shannon} = \sqrt{\left(D_{KL}(p||\frac{p+q}{2}) + D_{KL}(q||\frac{p+q}{2})\right)/2}$$

- **Common Part of Commuters (CPC)**: Researchers commonly utilize CPC to gauge the similarity among trajectories. In our study, we employ the Sørensen-Dice coefficient to evaluate the degree of similarity between a given trajectory and synthetic trajectories that share the same OD. For the given pair of trajectories, $\tau_{D_i}$ and $\tau_{P_i}$, their CPC is defined as follows:

$$Sørensen - Dice(\tau_{D_i}, \tau_{P_i}) = \frac{2|\tau_{D_i} \cap \tau_{P_i}|}{|\tau_{D_i} + \tau_{P_i}|}$$

### 3.4 Explainability of Learned Environmental Preference

We adopt XAI to address the challenge of explainability within AI models. As a representative method for XAI, SHAP (SHapley Additive exPlanations) has been extensively employed in studies such as trajectory synthesis and trajectory prediction (Lundberg & Lee, 2017; Simini et al., 2021). Specifically, Shapley offers a method based on game theory and local explanation to estimate the contribution of each feature value. In this method, the contribution of each feature value to the model output is allocated based on its marginal contribution as follows:

$$\phi_i(v) = \sum_{S \subseteq N \setminus \{i\}} \frac{|S|!(n-|S|-1)!}{n!}(v(S \cup i) - v(S))$$

where $\phi_i(v)$ refers to the contribution of feature $i$ to the model output $v$, $S$ represents the target sample, and $n$ represents the number of sample features. In the equation above, the first term can be understood as possible permutations, while the second term can be understood as the marginal effects of the corresponding permutations on the model output.

As an implementation approach of Shapley values in engineering, SHAP provides a feasible way to overcome high computational costs and reveal the contributions of explanatory variables in AI models. This study employs the SHAP method to interpret location preference derived from MEDIRL, with the goal of understanding the complex mechanisms underlying how built environments influence cycling decision-making behaviors.

# 4 Case Study

## 4.1 Descriptive Statistics of Model Output

We quantify cyclists' general street visual preferences by training the MEDIRL model in Bantian Sub-district in Longgang District, Shenzhen. Specifically, using an MDP with a 100m grid size and a 0.99 discount rate $\gamma$, our approach employs a Multi-Layer Perceptron (MLP) with 4 hidden layers and Rectified Linear Units (ReLU) to approximate the relationship between state features and the reward function, and the value of reward function at each state represents quantified street visual preferences.

Globally, Bantian Sub-district exhibits a mean normalized reward of 0.40 with a variance of 0.12. The overall distribution in this area shows a right-skewed pattern where the mode of quantified preference is less than its mean, as illustrated in **Fig. 6**(a). This indicates that Biantian Sub-district has many locations with relatively low reward values, indicating distinct preferences among riders. We visualize the spatial distribution of quantified cyclists' preferences, as shown in **Fig. 6**(b). The varying shades of basic study units represent the degree of preference that cyclists have for corresponding locations. The figure clearly indicates significant spatial clustering in cyclists' preferences, with less favored locations often situated along the edges between two communities.

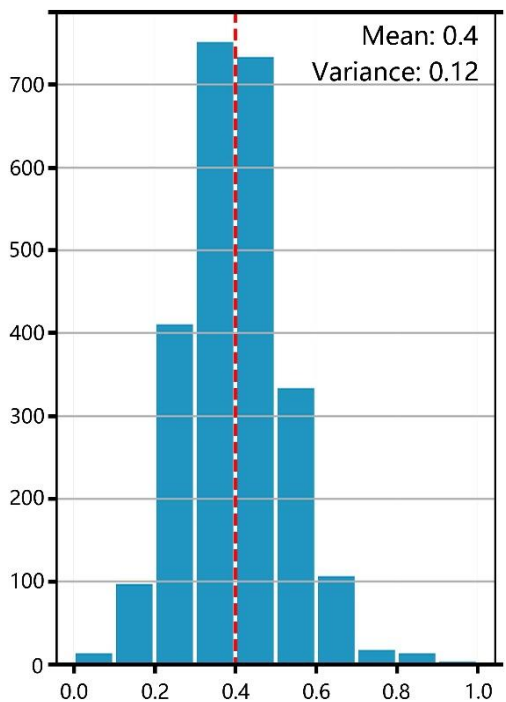
(a) Distribution of normalized cyclists' preferences

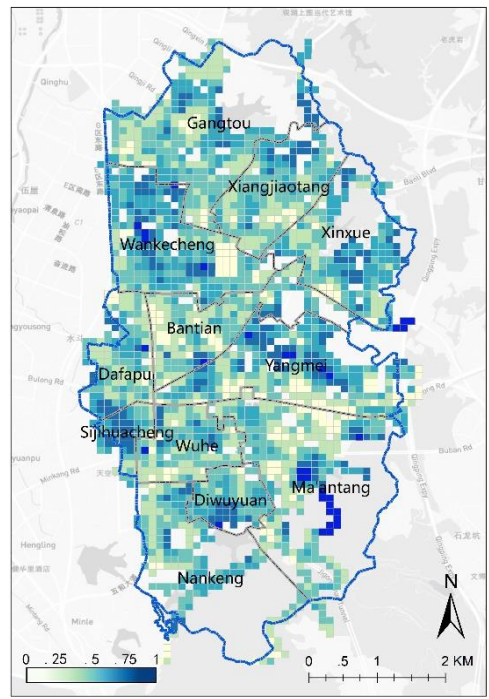
(b) Spatial distribution of normalized cyclists' preferences

Fig. 6. Visualization of cyclist' preferences

## 4.2 Model Evaluation

We assess the similarity between actual and synthetic trajectories based on their statistical characteristics as summarized in **Fig. 7**. The JSD between the real trajectories and the generated trajectories is 0.3484, indicating a significant degree of resemblance in terms of fluctuation ranges. Our model result performs better than existing research in similar scenarios (Zhao & Liang, 2023).

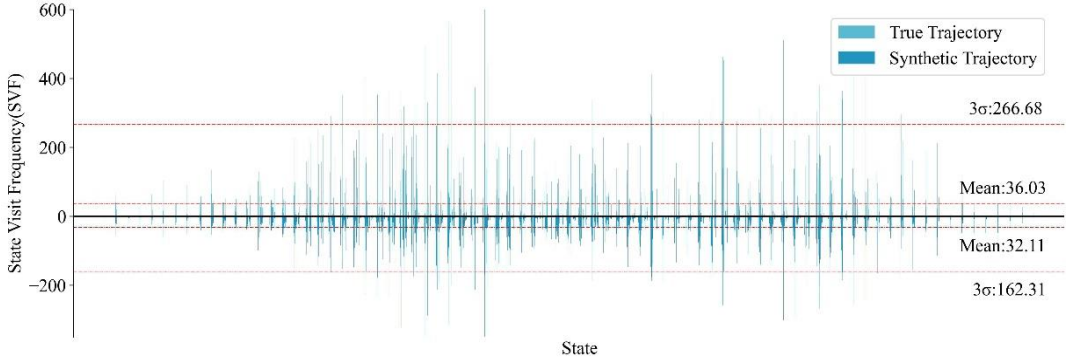

Fig. 7. SVF distributions of real and synthetic trajectories

Moreover, we utilize the Common Part of Commuters (CPC) to assess the similarity between individual trajectories (Cao et al., 2024; Liu et al., 2020; Liu & Jiang, 2022; Simini et al., 2021). Our findings show that, on average, synthetic trajectories derived from learned street visual preferences overlap with real trajectories by 73.77% for each OD pair, whereas the shortest paths between the same ODs overlap by 63.78%. As illustrated in **Fig. 8**, the average CPC between synthetic and real trajectories decreases as cyclists' decision frequency increases in a single trip. Cyclists navigating fewer than 8 road segments in their trips have an average CPC over 80%, highlighting the significant impact of street visual preferences on their cycling procedures. However, as the number of road segments increases, the influence of environmental factors on cycling route selection diminishes. Additionally, the boxplot in **Fig. 8**(a) illustrates that, in certain high-decision-frequency scenarios, the synthetic trajectories generated by our model closely resemble real DBS trajectories.

We also compare the performance of our MEDIRL-based model with the Dijkstra algorithm (Dijkstra, 1959; Hagberg et al., 2008), examining how the CPC of synthetic trajectories and shortest paths varies with decision frequency during a single trip. The results are shown in **Fig. 8** (b). When cyclists navigate fewer than 6 road segments in a trip, the synthetic trajectories and shortest paths show significant similarity. However, as decision frequency increases, our model outperforms the Dijkstra algorithm. For decision frequencies exceeding 34, the correlation between synthetic data and actual trajectories sharply declines due to insufficient data for long-distance cycling. Our findings indicate that trajectories generated using the learned reward function effectively reflect preferences for medium to short-distance cycling paths, though the model still performs well in various route decision scenarios.

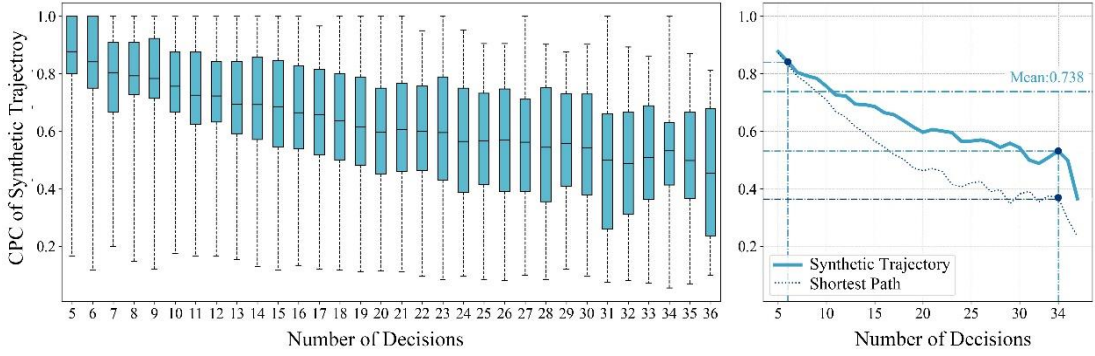

(a) Variation in synthetic data CPC with decision frequency during a single trip

(b) Variation in synthetic data and shortest path CPC

**Fig. 8. Variation in CPC with decision frequency during a single trip**

To gain more intuition about model outputs, we visualize examples of synthetic trajectories generated by our model alongside their corresponding shortest paths computed using the Dijkstra algorithm, and compare these with real trajectories, as shown in **Fig. 9**. Typically, the blue circles indicate the origin and destination and the sequence of black points represents the real GPS trajectory. The red curves in these figures represent the shortest paths. Meanwhile, the sequence of rectangles shows the reward learned using MEDIRL during the routes, while the yellow curves depict the synthetic trajectories based on this reward.

Overall, the trajectories generated by our model more closely resemble the real trajectories than their corresponding shortest paths, demonstrating that the MEDIRL model effectively captures cyclists' sequential decision-making patterns from real trajectory data. In **Fig.9**(a), the shortest path aligns with both the real and synthetic data, suggesting that the learned cycling rewards are consistent with the shortest path for some trips. In **Fig. 9**(c), although the shortest path closely matches the real trajectory, discrepancies indicate that cyclists make route decisions based on cycling rewards rather than just distance. In **Fig. 9**(b) and (d), the significant differences between the shortest paths and the synthetic data imply the importance of the learned rewards in the cycling procedures. Cyclists tend to select routes that offer higher cumulative rewards instead of the shortest path, even when a higher-reward state is close to their current position. This suggests that our model captures the nature of cycling route decision, and our quantified street visual preferences play a crucial role in cycling, given their high explanatory power in generating DBS data.

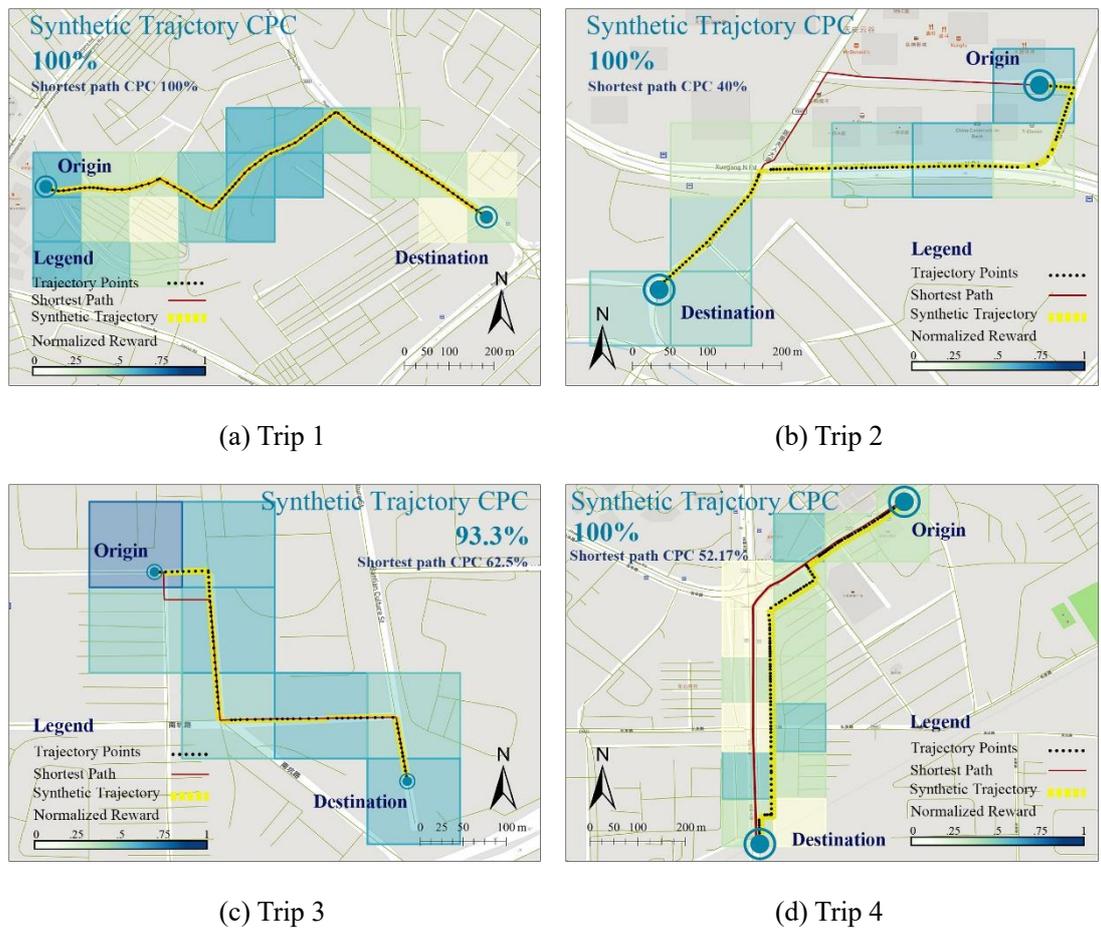

(a) Trip 1  (b) Trip 2

(c) Trip 3  (d) Trip 4

Fig. 9. Selected real and synthetic trajectories

### 4.3 Interpretability of Environmental Preference of Route Decision Process

#### 4.3.1 Importance of the Street Visual Elements

Our discussion centers on interpreting the MEDIRL reward function to discover interesting patterns. We display the distribution of SHAP values for street-level environmental elements, organized into primary categories, in **Fig. 10**. These factors display significant variations in their influence on cycling rewards, reflecting cyclists' diverse street visual preferences. Street visual elements related to vehicle, flat, construction and nature have a notable impact on cycling rewards. Specifically, the proportion of motorcycles in SVI has a much higher average SHAP value compared to other features. Following closely in contributions include proportion of wall, Sky View Ratio (SVR), proportion of fences, Green View Index (GVI), and proportion of terrain elements in cyclists' visual fields.

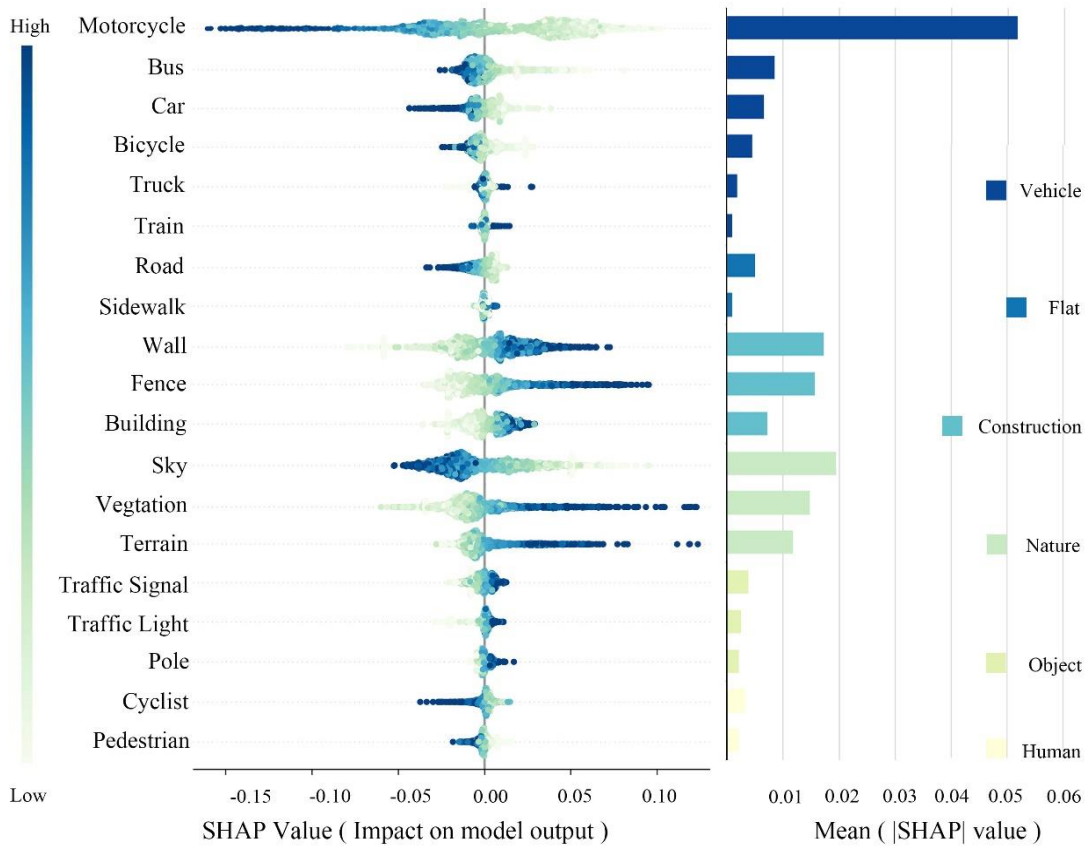

**Fig. 10. Distribution of Shapley values for all features in the discriminator of MEDIRL**

### 4.3.2 Impacts of Street Visual Elements on Cycling Preferences

We categorize cyclists' attention to visual elements into three primary domains based on their importance to previously quantified cycling preferences: safety, street enclosure and cycling comfort. First, cyclists are notably aware of their interaction with other transportation modes (Ito & Biljecki, 2021), highlighting their focus on safety while riding. Research emphasizes that measures reducing negative interactions and protecting cyclists' right of way, such as dedicated bicycle lanes, positively impact safety (Jeon & Woo, 2024; Kroesen & van Wee, 2022; Meng & Zheng, 2023; Wang et al., 2024). In our study, a low proportion of motorcycles and cars in their visual fields is assigned with a positive SHAP value, indicating that motor vehicles negatively affect their preferences. To put it differently, cyclists tend to choose routes with fewer motorcycles and cars, underscoring their prioritization of safety. Second, the impact of enclosure—measured by the proportion of vertical elements in SVI, such as walls, buildings, and fences—demonstrates a strong connection to bikeability and urban vitality, as noted by previous researchers (Ito & Biljecki, 2021; Meng & Zheng, 2023; Wang et al., 2019). Specifically, a greater presence of these elements in cyclists' visual fields is positively associated with their preferences, suggesting that these features can slightly increase the likelihood of selecting certain streets. Third, cycling comfort, represented by the GVI and SVI, serves as a crucial foundation for cycling preferences. Generally, cyclists prefer routes with higher GVI for enhanced comfort, while those with elevated SVR are less appealing, indicating a preference for shaded areas over open skies, which is especially important in subtropical regions. However, there are no significant relationships between infrastructure elements, such as the proportion of traffic lights and poles in cyclists' visual fields, as confirmed by related research (Ito & Biljecki, 2021). It can be elucidated that these factors do not exhibit significant variations in our research area and do not contribute to differences in quantified street visual preferences among cyclists.

The preceding discussion underscores the diverse impacts of street visual elements on cycling quantified by MEDIRL, revealing their multifaceted contributions to cyclists' preferences. However, we find that the same street visual elements have varying impacts across different states and trips. In other words, the complex relationship between street visual elements and cycling procedure remains unclear. This ambiguity hinders the development of cost-effective strategies for designing streetscapes and improving cycling experiences. To explore this further, we identify six elements with high global importance based on expected SHAP values. We draw their local dependency plots on cycling rewards in **Fig. 11**. These plots illustrate how changes in each element affect overall cycling preferences: higher y-axis values indicate greater impact in reward explanation, while steeper slopes indicate marginal effects to those changes (Lundberg & Lee, 2017).

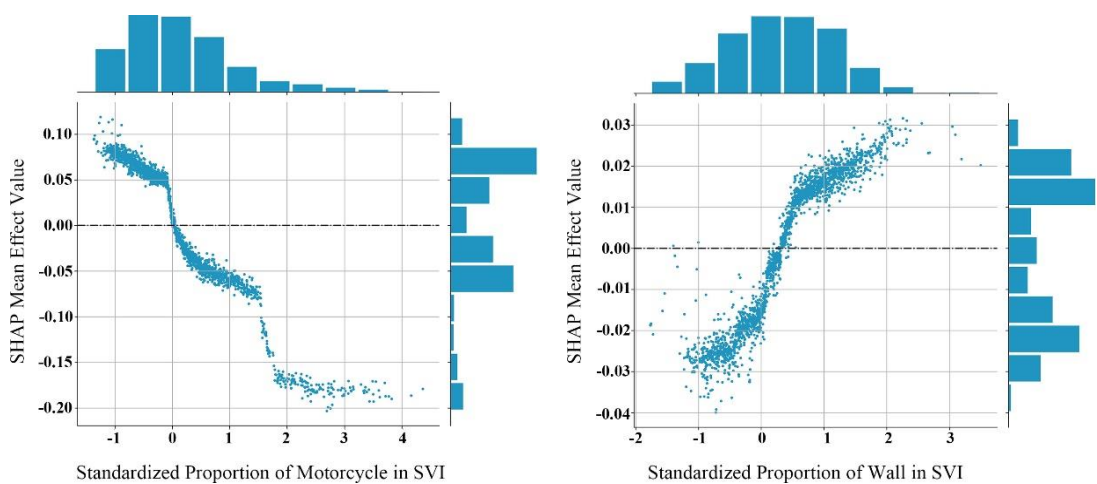

(a) Dependence plot of motorcycle proportion   (b) Dependence plot of wall proportion

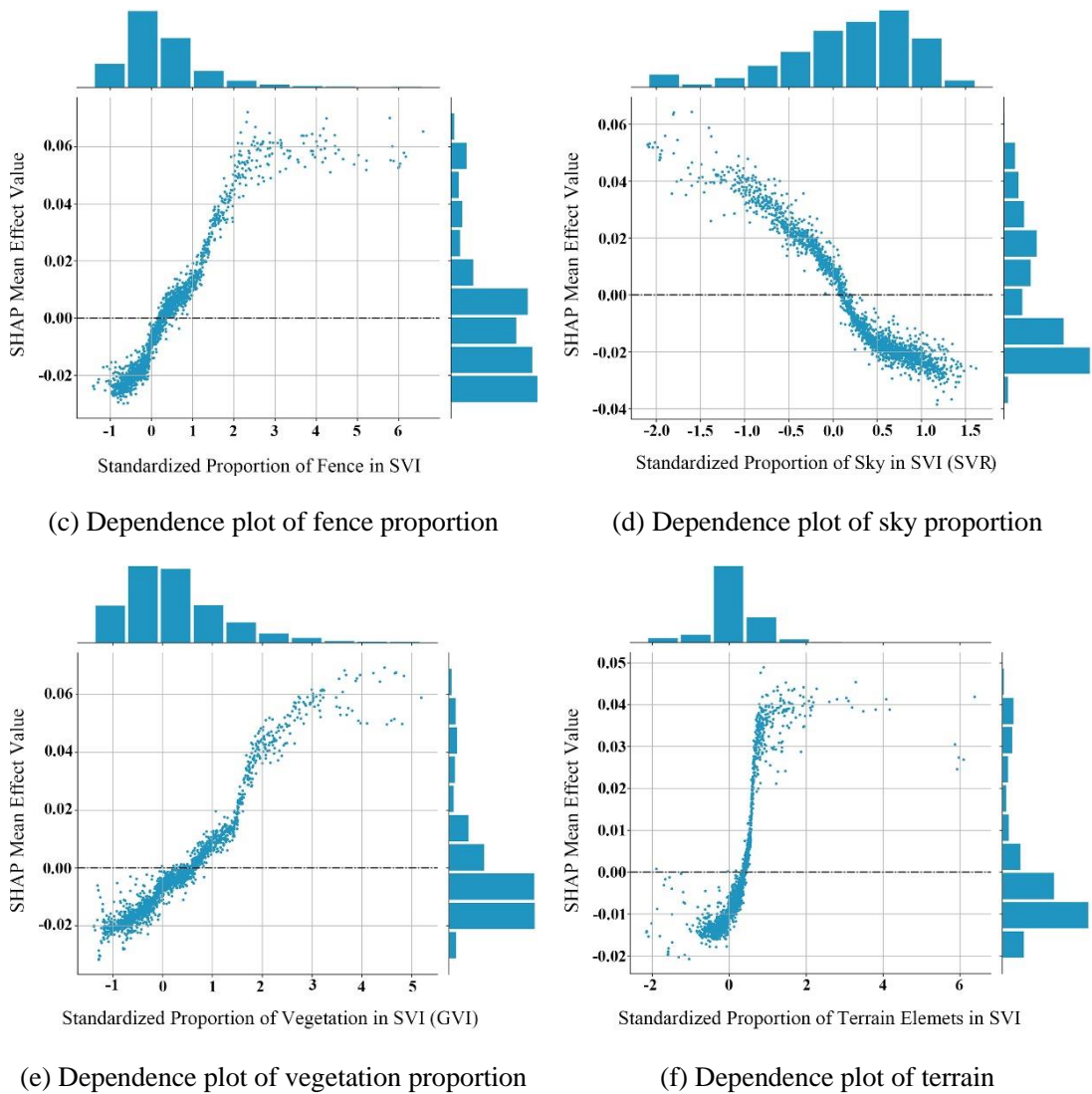

(c) Dependence plot of fence proportion   (d) Dependence plot of sky proportion

(e) Dependence plot of vegetation proportion   (f) Dependence plot of terrain

**Fig. 11. Dependence plot of key street visual elements**

The findings demonstrate that the selected street visual elements do not fit into simple categories of positive or negative impacts. Instead, their effects vary depending on their change of proportions. Specifically, the proportion of motorcycles in cyclists' visual fields has a contribution to their route choices that initially promotes and then inhibits as demonstrated in **Fig.11**(a). This supports prior research on the negative interaction between cyclists and vehicles on DBS volume (Ito & Biljecki, 2021; Jeon & Woo, 2024; Mertens et al., 2016). Additionally, the marginal effects of motorcycle SHAP values reveal a roughly four-stage pattern: cyclists generally prefer states with lower traffic volume and avoid those with higher traffic volume. Within the standardized range of 1.5 to 2, cyclists are highly sensitive to changes in the number of motorcycles. This sensitivity quickly levels off when the standardized ratio approaches 4.0, indicating the most cost-effective zone for traffic calming.

We further examine the impact of key elements related to street enclosure on cycling procedure across different states. Overall, the contribution of walls and fences to cycling preferences exhibits a distinct two-stage trend shown in **Fig.11**(b) and (c). The proportion of walls shows a negative impact when standardized measures are below 0.5, indicating cyclists tend to avoid streets with minimal enclosure. This trend reverses around and beyond a value of 1. Similarly, the proportion of fences initially has a negative impact but then gradually exerts increasingly positive effects on quantified cycling preferences. We also discover threshold effects within the variation of SHAP values. The SHAP value for walls shows a diminishing marginal effect around 0.5, indicating a threshold which additional wall proportions do not significantly improve cycling rewards. In addition, the SHAP value for fences exhibits a two-stage trend, stabilizing with a plateaued standardized SHAP value around 2.0. This suggests that the presence or absence of fences, rather than their quantity, markedly influences cyclists' route choices, which goes beyond the analysis of existing studies (Meng & Zheng, 2023; Song et al., 2024; Wang et al., 2019).

We also explore the SHAP variation patterns for factors related to cycling comfort as shown in **Fig.11**(d), (e) and (f). In **Fig.11**(d), the contribution of SVR promotes and then inhibits cycling comfort, with its SHAP value reflecting a similar two-stage pattern. In **Fig.11**(e), GVI demonstrates an opposite trend compared to the SVR. Initially, vegetation proportion in cyclists' visual fields negatively impacts their preferences, but as the standardized GVI increases, SHAP values stabilize, and in some cases, decrease. This might be due to high GVI being associated with a lack of necessary public facilities. This variation helps explain conflicting views on the impact of street greenery (Blitz, 2021; Song et al., 2024). Additionally, the effect of terrain on cycling preference shows an inverted U-shape trend, initially promoting but diminishing later on. In **Fig.11**(f), it indicates that the impact of horizontal greenery is greatest when the standardized value is approximately 2.0, corresponding to a true value of about 8%.

In summary, street visual environmental elements exhibit complex relationships with cycling behaviors, often demonstrating threshold effects. Specifically, changes in the proportion of motorcycles illustrate a nonlinear impact on cyclist preferences regarding safety. Visual elements that enhance visual enclosure —such as the proportion of walls and fences —and those improving comfort, like SVR and GVI, show marginal effects of their SHAP values gradually decreasing and eventually converging. Therefore, strategically managing the proportions of visual elements on streets is crucial for influencing cyclists' route choices, promoting cycling activities, and encouraging favorable cycling behaviors.

### 4.3.3 Interpretation of Selected Trips Using Street Visual Elements

To better understand the environmental effects on cycling behavior, we select several records from previously analyzed trajectories, examine their key states, and investigate the local effects of street visual elements on cycling rewards, as illustrated in **Fig.12**. Similar to the previous analysis, the blue circles indicate the origins and destinations, while the sequence of rectangles displays the rewards learned using MEDIRL throughout the routes. Additionally, the black points represent the locations of selected SVIs, with key states in the selected trips highlighted in black boxes.

In the analysis of key states in selected trip 1, as shown in **Fig.12**(a). We observe significant spatial heterogeneity in cyclists' preferences for specific street visual elements along the route. Generally, a higher value of reward function correlates with a high SHAP value for the proportion of motorcycles, while elements related to street enclosure and cycling comfort have opposite effects at the beginning and end of the trip. In the analysis of trip 2 (**Fig.12**(b)), we find that cyclists are more inclined to select the current route over alternatives when key states indicate higher quantified cycling preference. This preference is linked to the significant contribution of the GVI and the proportion of motorcycles in their visual fields. This suggests that the characteristics of the streetscapes along the livelihood streets support cyclists' sense of safety and enhance their comfort while cycling. In selected trip 3, illustrated in **Fig.12**(c), the streetscapes along the route are characterized by a modern urban landscape, with street visual elements related to enclosure—such as fences, walls, and buildings—playing a crucial role in shaping cyclists' preferences. Moreover, a higher proportion of sky negatively affects cyclists' tendency of selecting corresponding roads. The quantified cycling preferences observed in selected trip 4 (**Fig.12**(d)) resemble those in selected trip 3. The overall reward is relatively lower than other areas. The proportion of motorcycles in cyclists' visual fields negatively impacts their route decision, and a high SVR and low GVI significantly hinder cyclists from choosing the corresponding roads.

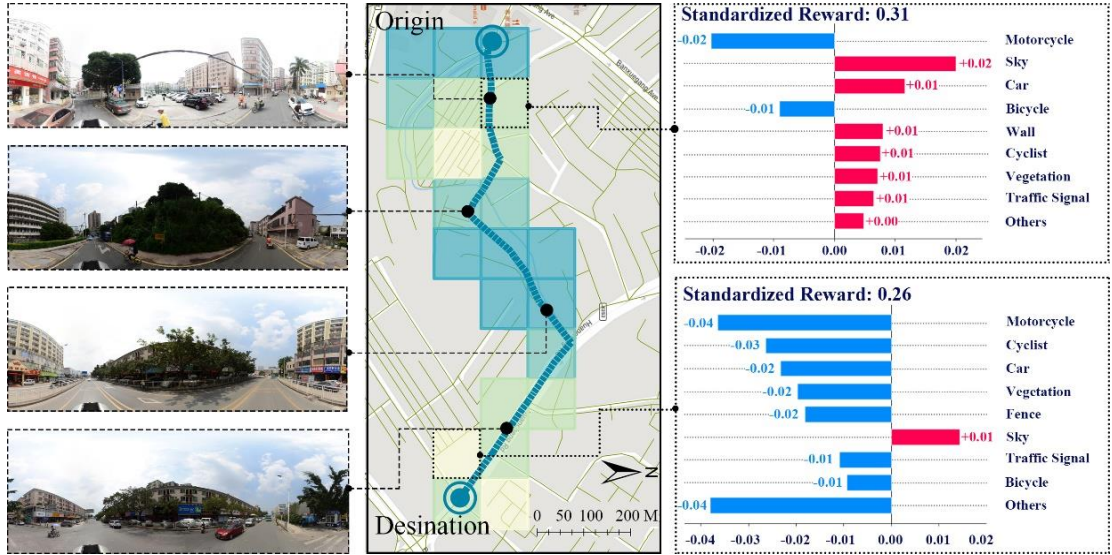

Streetscapes along the route     Trajectory of trip 1     Local interpretation of key states

**(a) Detailed cycling procedure of selected trip1**

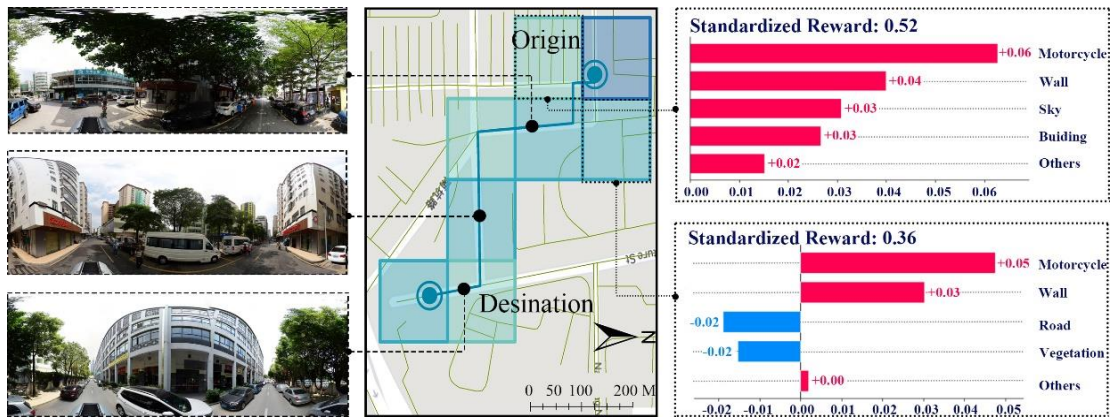

Streetscapes along the route     Trajectory of trip 2     Local interpretation of key states

**(b) Detailed procedure of selected trip 2**

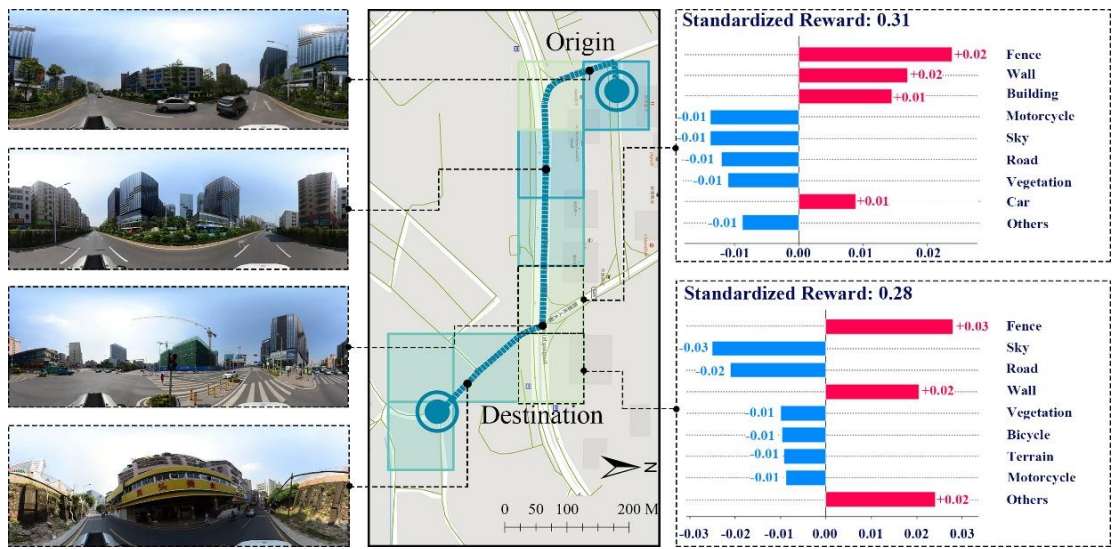

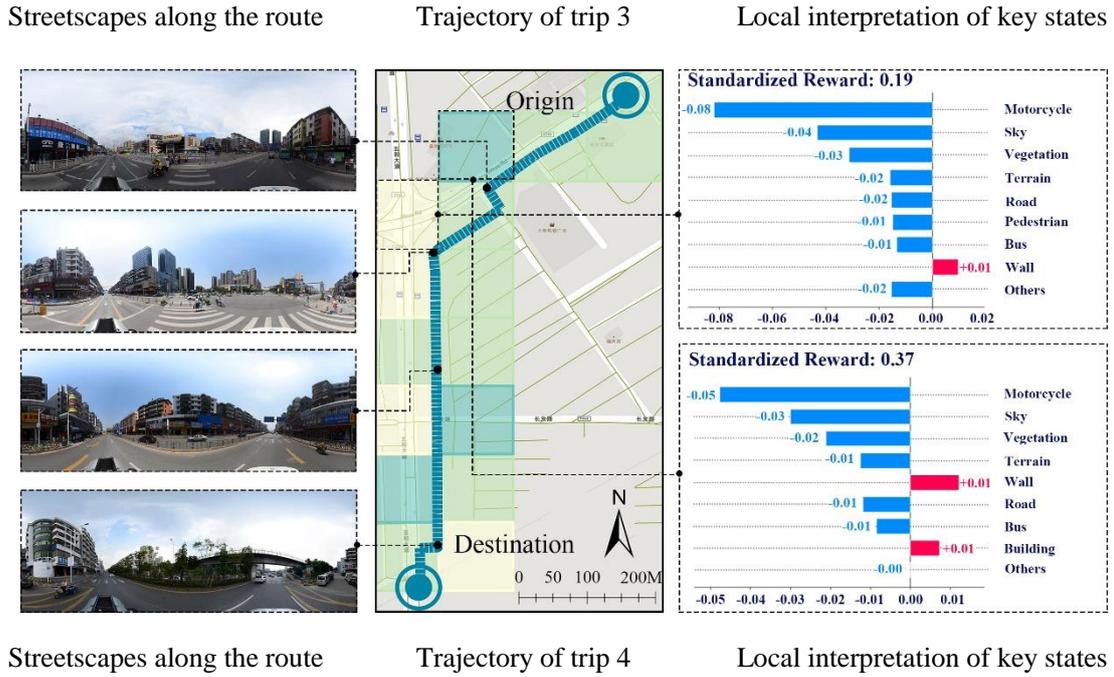

Fig.12 Detailed procedure of selected trips

## 5 Conclusion and Discussion

Our study automatically quantifies and interprets cyclists' complicated street visual preferences from cycling records by leveraging MEDIRL and XAI in Bantian Sub-district of Longgang District, Shenzhen. Specifically, we formulate the cycling procedure as an MDP and further employ MEDIRL to recover the underlying reward function in the context of RCM problems, which serves as cyclists' preferences as they travel between their initial positions to destinations based on their real-time street visual perceptions. Additionally, we utilize XAI techniques to discern the characteristics of streetscape that cyclists favored.

The major contributions of this study can be concluded in three aspects:

(1) We proposed a comprehensive framework for unraveling the cycling preference of urban streetscapes, emphasizing a detailed procedure that goes beyond conventional OD analysis. DBS trajectories offer insights into the entire cycling process, enabling a more nuanced understanding of RCM, while SVI provides opportunities to accurately describe street-level visual features related to cycling trips.

(2) We proposed a novel cycling preference quantification and interpretation method, leveraging results derived from MEDIRL and XAI models. This method automatically establishes a robust link between the street environment and cycling behavior, explicitly considering their complex relationships. Compared to existing methods, our data-driven method offers enhanced reliability and reasonability.

(3) We applied our proposed framework in a real-world scenario, specifically in Bantian Sub-district in Longgang District, Shenzhen. The results enrich the current research by considering the complex nonlinear effects of the street visual environment on cycling from various dimensions. Our findings provide valuable practical insights into bicycle-friendly streetscape design.

The results indicate that, firstly, the MEDIRL can effectively infer the underlying behavioral principal of the cycling process, which can be interpreted as preferences. The synthetic trajectories guided by our inferred preferences can well replicate the statistical and path characteristics of actual trajectories. Secondly, our study examines the learned cyclist preferences and their relationships with street visual environmental factors. We find that cyclists are particularly concerned with specific street visual elements, reflecting their focus on safety and their preference for links characterized by high enclosure and a comfortable environment. Thirdly, cyclists' attention to specific street visual elements display nonlinear characteristics and threshold effects. Urban planners can improve cycling experience by addressing cyclists' needs at the micro-level of street design. Typically, implementing traffic calming measures to enhance safety and improve street bikeability are available (Ito & Biljecki, 2021; Meng & Zheng, 2023; Wang et al., 2024; Winters et al., 2013). Enhancing street enclosure and continuity in a cost-effective manner, considering the threshold effect of environmental elements, is also beneficial.

While our framework can be applied to other areas to understand visual preferences, it has limitations. In this study, we regard cycling as continuous route choice procedures influenced by streetscape characteristics, which helps identify general preferences. However, individual travel patterns can affect these preferences, showing that different cyclists may have different needs. Future research should focus on improving how trajectory data is represented and including more semantic details about the routes. This would not only increase the comparability of different trajectory data but also provide theoretical foundations and technical means to explore the similarities and differences in patterns of route decision among different types of cyclists.

In summary, our findings help to discover cyclists' general street visual preferences based on their continuous route decision procedures influenced by streetscape characteristics. Additionally, the intuitively understandable insight supports urban planners in designing and updating streetscapes to meet cyclists' specific needs and promote cycling.

# 6 Appendix

## 6.1 Time differentiation characteristics of DBS

To explore the differentiation of DBS cycling behaviors across different time periods, we calculate the Manhattan distances for different travel trips. Due to the log-normal distribution characteristics of cycling distances in statistics, we plot the frequency distribution histogram of the logarithm of cycling distances as shown in **Fig. 14**. The results indicate that the logarithmic mean of cycling distances on workdays is approximately 2.3, whereas on weekends, the mean logarithm of travel distances is 2.4. Therefore, compared to workdays, cyclists travel longer distances on weekends, and the variance from the distribution curve also shows greater fluctuations in weekend travel distances. Similarly, the logarithmic mean of nighttime cycling distances is 2.0, which rises to 2.3 during the daytime. Thus, compared to daytime, nighttime cyclists travel shorter average distances with greater fluctuations. Overall, these results align with researchers' intuitive understanding of DBS, indicating that cyclists' behavioral patterns exhibit certain differences across time periods.

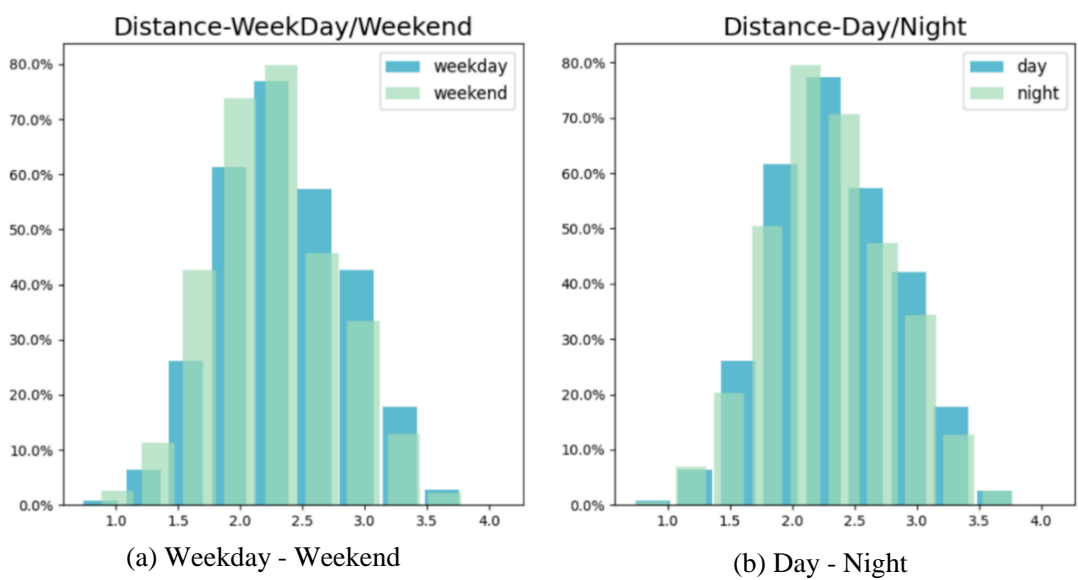

(a) Weekday - Weekend  (b) Day - Night

**Fig. 14. Variation of cycling distance on time period**

Our study investigates the nuanced characteristics of riding distances at a granular scale. **Fig. 15** illustrates the logarithm of DBS travel distances on the x axis against travel time on the y axis. For clarity, we aggregate the number of DBS trips per hour, depicted in the subplot on the right. Overall, as indicated by the red dashed line in the figure, the average logarithmic travel distance is 2.4. Early morning trips often deviate significantly from this average, while daytime trips tend to align more closely. However, around noon, distances diverge further from the mean. Additionally, travel distances on weekends frequently differ from weekdays. This data distribution highlights substantial variations in cyclists' behavior influenced by daily rhythms. Notably, DBS peak in usage on weekdays from 6:00-9:00 AM and 4:00-8:00 PM. Meanwhile, trips in the early morning of weekends are notably fewer. The variability in travel distances follow distinct statistical patterns throughout different times of the day.

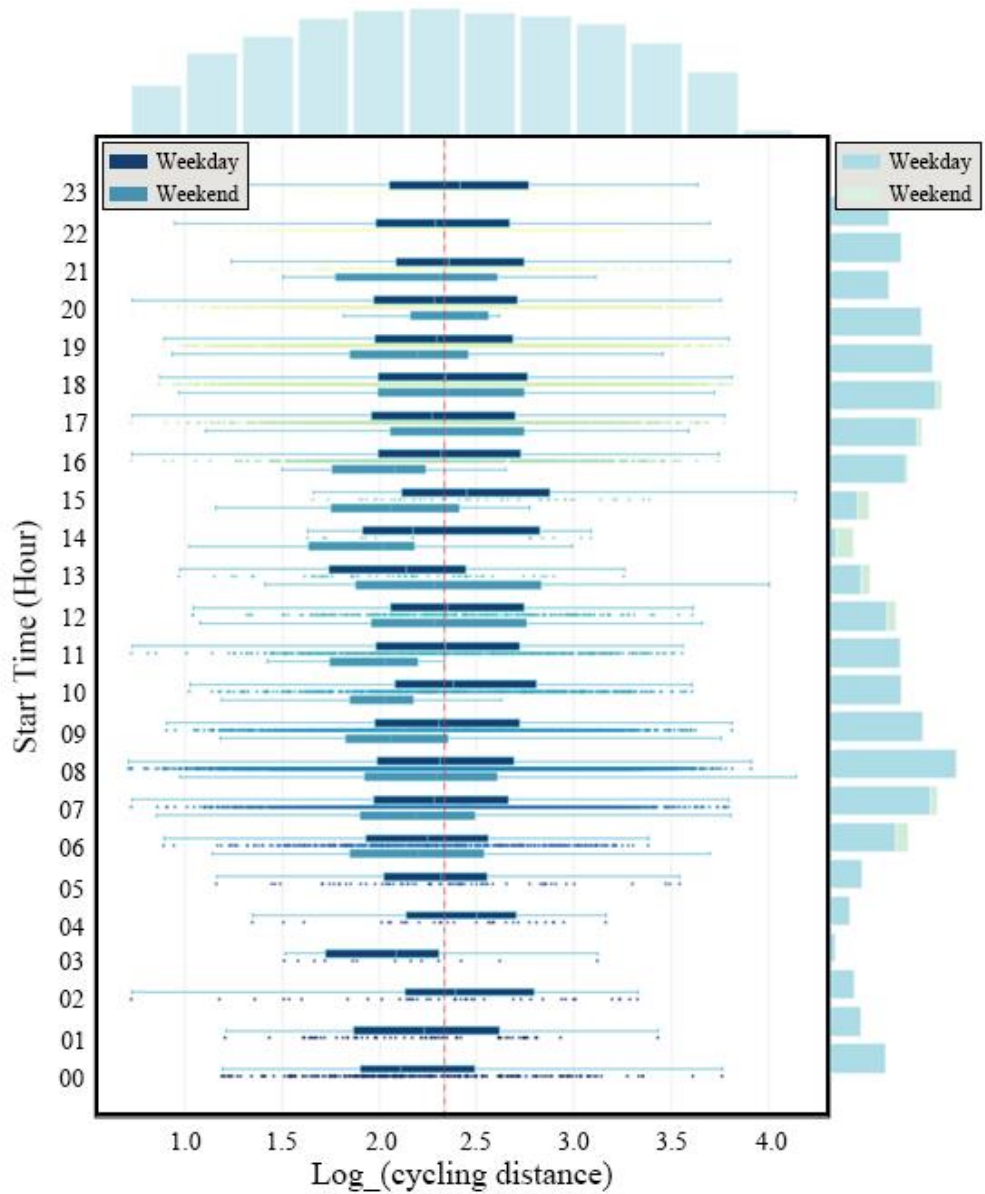

**Fig. 15. Hourly variation of cycling distance**

## 6.2 The difference between DBS trajectory and corresponding shortest path

Previous research has shown considerable variation between cyclists' actual route choices and the possible shortest paths. To assess the applicability of these findings in our context, we compare the similarity between the shortest paths and cyclists' actual trajectories. Firstly, we construct a directed spatial network based on roads in our research area, weighted by segment lengths. Secondly, we employ the Dijkstra algorithm (Dijkstra, 1959) on this road network to determine the shortest route for each cycling journey. Lastly, by using similarity metrics, our study contrasts the differences and similarities between cyclists' actual route choices and the shortest paths, thereby providing criteria for decision-making in data selection.

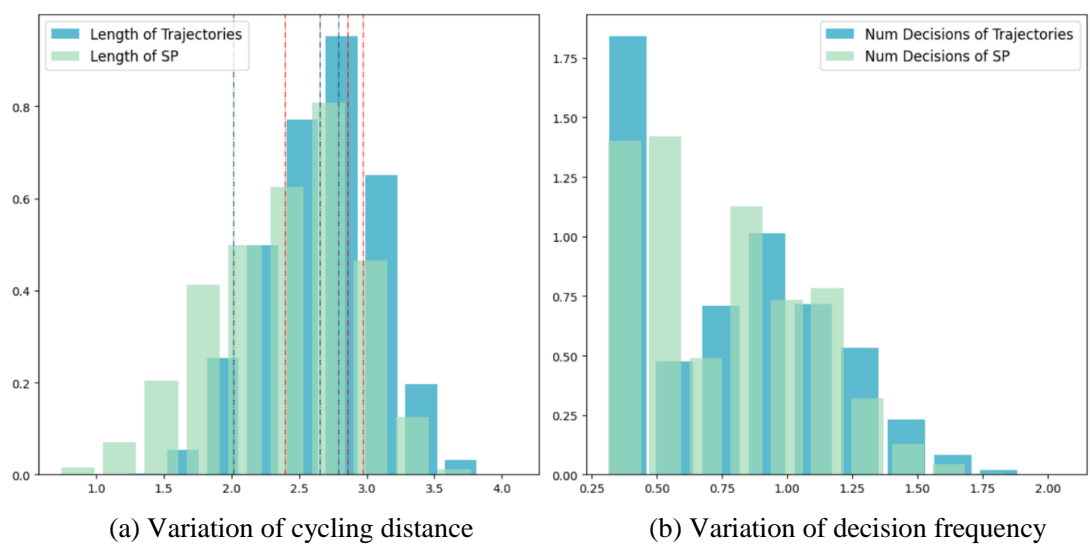

(a) Variation of cycling distance        (b) Variation of decision frequency

**Fig. 16. The similarities between trajectories and shortest path**

Initially, we analyze the similarity between cycling distance and the frequency of cycling decision-making, depicting their log-transformed frequency distribution histograms in **Fig. 16**. Regarding cycling distances, as shown in **Fig. 16**(a), after log transformation, both distributions exhibited varying degrees of left-skewed normal distribution, with more data concentrated to the left of the mean and some extreme values to the right. Specifically, the mean log-transformed actual trajectory cycling distance was 2.9, whereas the mean log-transformed shortest path cycling distance was 2.6. This suggests that cyclists do not strictly follow the shortest path when making route decisions. We also annotated the quartiles of both trajectories, indicating that the fluctuation in cycling distance for actual paths was smaller than that for the shortest path.

Additionally, we plotted the probability mass distribution of cycling decision-making frequencies with a single trip, as depicted in **Fig. 16**(**b**). Two peaks were evident around 3 and 8 decisions. This phenomenon indicates that as the decision frequency during a single trip increases during a single trip, cycling behavior shows different trends, highlighting the necessity for data filtering. By comparing the differences in number of decisions between actual trajectories and the shortest path, we observed greater similarity in their distributions around the first peak. This implies that cyclists' route decision patterns align more closely with the shortest path when fewer decisions are made in a single trip. However, around the second peak, differences emerged, with actual cycling trajectories exhibiting characteristics of high expectation, left-skewness, and long tails. In other words, within this range, cyclists' real route choices often deviate from the shortest path.